\documentclass{article}

\PassOptionsToPackage{numbers, compress, sort}{natbib}

\usepackage[final,nonatbib]{neurips_2021}

\usepackage[maxnames=12]{biblatex}

\usepackage{microtype}
\usepackage{graphicx}
\usepackage{subfigure}
\usepackage{booktabs} 
\usepackage{xcolor}
\usepackage[utf8]{inputenc} 
\usepackage[T1]{fontenc}    
\usepackage{hyperref}       
\usepackage{url}            
\usepackage{booktabs}       
\usepackage{amsfonts}       
\usepackage{amsmath}
\usepackage{mathtools}
\usepackage{amsthm}
\usepackage{amssymb}
\usepackage{nicefrac}       
\usepackage{microtype}      
\usepackage{multirow}
\usepackage{colortbl}
\usepackage{array}
\usepackage{booktabs}
\usepackage{algorithm}
\usepackage[noend]{algpseudocode}
\usepackage[capitalize]{cleveref}
\usepackage{xspace}

\usepackage{wrapfig}
\usepackage{caption}
\usepackage{soul}

\algrenewcommand\algorithmicrequire{\textbf{Input:}}
\algrenewcommand\algorithmicensure{\textbf{Output:}}

\newtheorem*{theorem*}{Theorem}

\DeclarePairedDelimiterX{\inp}[2]{\langle}{\rangle}{#1, #2}

\title{Opacus: User-Friendly Differential Privacy Library in PyTorch}

\author{%
  Ashkan Yousefpour\thanks{Equal contribution. Jessica Zhao contributed benchmarking code and analysis (Section~\ref{sec:experiments}). Correspondence to \texttt{yousefpour@fb.com}.}\And Igor Shilov\footnotemark[1]\And Alex Sablayrolles\footnotemark[1]\And Davide Testuggine\And Karthik Prasad\And Mani Malek\And John Nguyen\And Sayan Ghosh\AND Akash Bharadwaj\And Jessica Zhao\footnotemark[1]\And Graham Cormode\And Ilya Mironov\AND
  \textnormal{Meta AI}
}

\begin{document}

\maketitle

\begin{abstract}
We introduce Opacus, a free, open-source PyTorch library for training deep learning models with differential privacy (hosted at \href{https://opacus.ai}{\texttt{opacus.ai}}). 
Opacus is designed for simplicity, flexibility, and speed. 
It provides a simple and user-friendly API, and enables machine learning practitioners to make a training pipeline private by adding as little as two lines to their code. 
It supports a wide variety of layers, including multi-head attention, convolution, LSTM, GRU (and generic RNN), and embedding, right out of the box and provides the means for supporting other user-defined layers. 
Opacus computes batched per-sample gradients, providing higher efficiency compared to the traditional ``micro batch'' approach. 
In this paper we present Opacus, detail the principles that drove its implementation and unique features, and benchmark it against other frameworks for training models with differential privacy as well as standard PyTorch. 
\end{abstract}

\section{Background and Introduction}
\label{sec:intro}
Differential privacy (DP)~\cite{dwork2006calibrating} has emerged as the leading notion of privacy for statistical analyses.
It allows performing complex computations over large datasets while limiting disclosure of information about individual data points.
Roughly stated, an algorithm that satisfies DP ensures that no individual sample in a database can have a significant impact on the output of the algorithm, quantified by the privacy parameters $\epsilon$ and $\delta$.

\begin{figure}[h]
\includegraphics[width=0.9\textwidth]{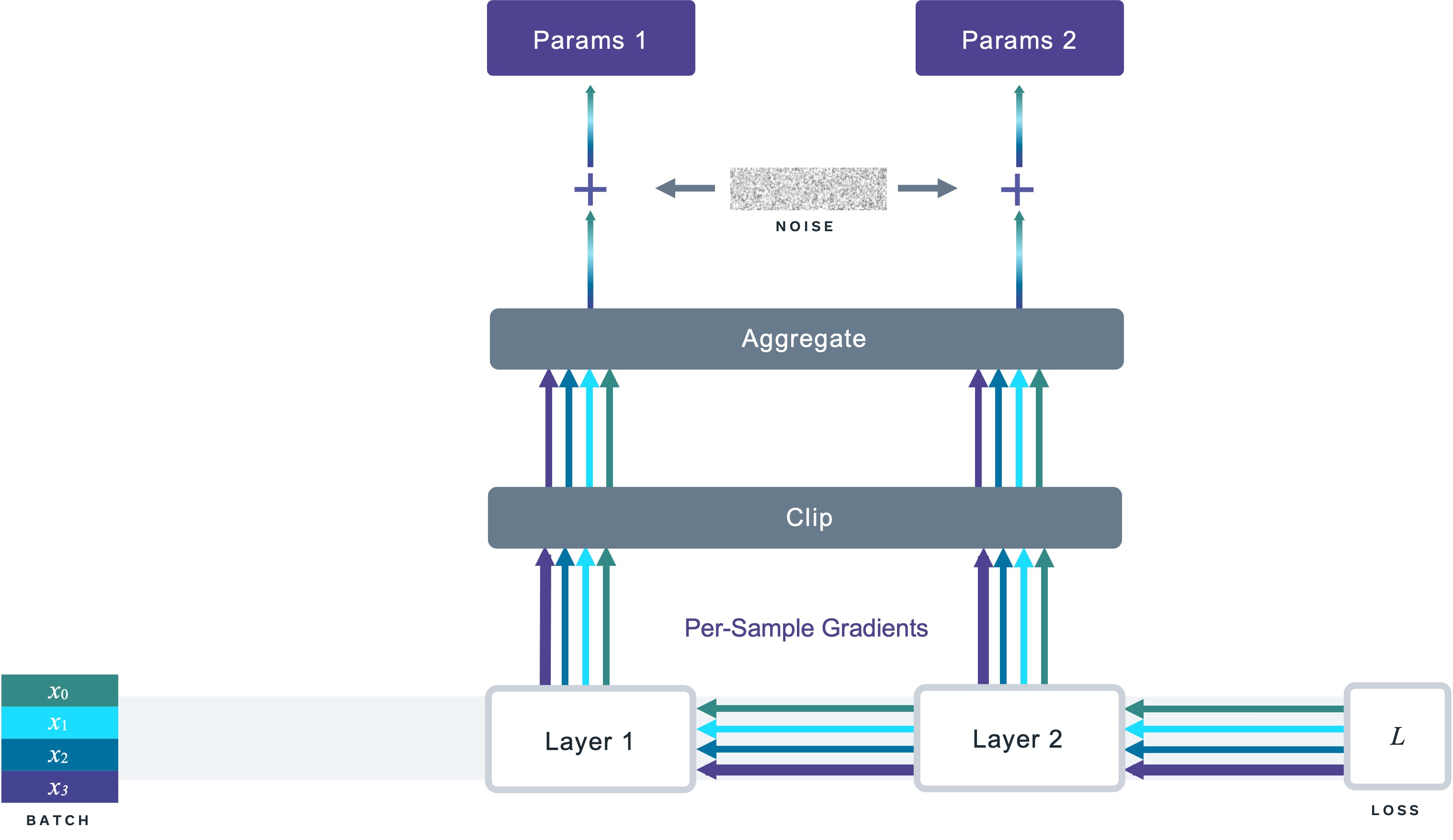}
\centering
\caption{Pictorial representation of the DP-SGD algorithm. The single-colored lines represent per-sample gradients, the width of the lines represent their respective norms, and the multi-colored lines represent the aggregated gradients.}
\label{fig:dp-sgd-image}
\end{figure}

Formally, a randomized mechanism $M\colon \mathcal{D} \rightarrow \mathcal{R}$ is $(\epsilon, \delta)$-differentially private for $\epsilon>0$ and $\delta\in[0, 1)$ if for any two neighboring datasets $D, D'\in\mathcal{D}$ (i.e., datasets that differ in at most one sample) and for \textit{any} subset of outputs $R\subseteq\mathcal{R}$ it holds that
\begin{equation*}
    \mathbb{P}(M(D) \in R) \leq \exp( \epsilon ) ~ \mathbb{P}(M(D') \in R) + \delta.
\end{equation*}
Differentially Private Stochastic Gradient Descent (DP-SGD) due to Abadi et al.~\cite{Abadi_2016}, building on Song et al.~\cite{song2013stochastic} and Bassily et al.~\cite{bassily2014private}, is a modification of SGD that ensures differential privacy on every model parameters update. Instead of computing the average of gradients over a batch of samples, a DP-SGD implementation computes per-sample gradients, clips their $\ell_2$ norm, aggregates them into a batch gradient, and adds Gaussian noise (see \cref{fig:dp-sgd-image} for an illustration.)
However, mainly for efficiency reasons, deep learning frameworks such as PyTorch or TensorFlow do not expose intermediate computations, including per-sample gradients; users only have access to the gradients averaged over a batch.

A na\"ive way to implement DP-SGD is thus to separate each batch into micro-batches of size one, compute the gradients on these micro-batches, clip, and add noise (see \cref{sec:microbatching} for sample code to obtain the per-sample gradients using this approach). 
While this procedure (called the ``micro-batch method'' or ``micro-batching'') does indeed yield correct per-sample gradients, it can be very slow in practice due to underutilization of hardware accelerators (GPUs and TPUs) that are optimized for batched, data-parallel computations.

Opacus implements performance-improving vectorized computation instead of micro-batching. 
In addition to speed, Opacus is designed to offer simplicity and flexibility. 
In this paper, we discuss these design principles, highlight some unique features of Opacus, and evaluate its performance in comparison with other DP-SGD frameworks. 

\section{Design Principles and Features}
\label{sec:design}
Opacus is designed with the following three principles in mind: 

\begin{itemize}
    \item {\em Simplicity}: Opacus exposes a compact API that is easy to use out of the box for researchers and engineers. Users need not know the details of DP-SGD in order to train their models with differential privacy. 
    \item {\em Flexibility}: Opacus supports rapid prototyping by users proficient in PyTorch and Python, thanks to its rich set of features (described below).
    \item {\em Speed}: Opacus seeks to minimize performance overhead of DP-SGD by supporting vectorized computation.
\end{itemize}

We explain throughout the paper how these principles manifest themselves in the Opacus API.

\paragraph{Example Usage.}
The main entry point to Opacus is the \texttt{PrivacyEngine} class. It keeps track of ``privacy budget'' spent so far and is responsible for wrapping regular PyTorch training objects with DP-related code. The key method provided by \texttt{PrivacyEngine} is \texttt{make\_private()}. It takes the three PyTorch training objects---model, optimizer and data loader--- along with the privacy parameters (noise multiplier and maximum norm of the gradients) and outputs differentially private analogues of these objects:
\begin{itemize}
    \item the model wrapped with \texttt{GradSampleModule}, which adds the ability to compute per-sample gradients;
     \item the optimizer wrapped with an additional code for clipping gradients and adding noise;
    \item the data loader transformed into one using Poisson sampling as required by DP-SGD.
\end{itemize}

This design provides a good balance between {\em simplicity} and {\em flexibility}. On the one hand, most users are able to switch to DP training by calling a single method. On the other hand, advanced users can modify the details of the private components' behavior as long as their interface is unchanged. 

Attaching Opacus to an existing script can be done with changing as few as two lines of code, e.g., the lines containing \texttt{privacy\_engine} in the following example:
{\small
\begin{verbatim}
   dataset = Dataset()
   model = Net() 
   optimizer = SGD(model.parameters(), lr) 
   data_loader = torch.utils.data.DataLoader(dataset, batch_size)
   privacy_engine = PrivacyEngine()
   model, optimizer, data_loader = privacy_engine.make_private(
      module=model,
      optimizer=optimizer,
      data_loader=data_loader,
      noise_multiplier=noise_multiplier,
      max_grad_norm=max_grad_norm,
   )
   # Now it's business as usual
\end{verbatim}}

\paragraph{Main features of Opacus.} We highlight some of the key features Opacus provides.

\textit{Privacy accounting.}
Opacus provides out of the box privacy tracking with an accountant based on Rényi Differential Privacy~\cite{mironov2017renyi, mironov2019r}. 
The \texttt{PrivacyEngine} object keeps track of how much privacy budget has been spent at any given point in time, enabling early stopping and real-time monitoring. Opacus also allows a user to directly instantiate a DP training targeting an  $(\epsilon, \delta)$ budget. In this instance, the engine computes a
noise level~$\sigma$ that yields an overall privacy budget of $(\epsilon, \delta)$. Opacus also exposes an interface to write custom privacy accountants.

\textit{Model validation.} 
Before training, Opacus validates that the model is compatible with DP-SGD. 
For example, certain layers (e.g., \texttt{BatchNorm} or \texttt{GroupNorm} modules in some configurations) mix information across samples of a batch, making it impossible to define a per-sample gradient; Opacus disallows those modules. 
It also ensures that no additional statistics without DP guarantees are tracked by the model (See \cref{sec:violation}). 

\textit{Poisson sampling.} 
Opacus also supports uniform sampling of batches (also called Poisson sampling): each data point is independently added to the batch with probability equal to the sampling rate.
Poisson sampling is necessary in some analyses of DP-SGD~\cite{mironov2019r}.

\textit{Efficiency.} Opacus makes efficient use of hardware accelerators (See~\cref{sec:vectorized}). Opacus also supports distributed training via PyTorch's \texttt{DistributedDataParallel}.

\textit{Virtual steps.}
As Opacus is highly optimized for batched per-sample gradient computation, it faces an inevitable speed/memory trade-off. In particular, when computing per-sample gradients for the entire batch, the size of the gradient tensor it needs to store is increased by the factor of \texttt{batch\textunderscore size}. To support a wider range of batch sizes, Opacus provides an option to decouple physical batch sizes, limited by the amount of memory available, and logical batch sizes, whose selection is driven by considerations of model convergence and privacy analysis.

\textit{Predefined and custom layers.}
Opacus comes with several predefined layer types, including convolution, multi-head attention, LSTM, GRU (and generic RNN), normalization, and embedding layers. 
Moreover, it allows users to add their own custom layers. When adding a custom layer, users can provide a method to calculate per-sample gradients for that layer and register it with a simple decorator provided by Opacus. The details can be found in \href{https://opacus.ai/tutorials}{\texttt{opacus.ai/tutorials}}. 

\textit{Secure random number generation.}
Opacus offers a cryptographically safe (but slower) pseudo-random number generator (CSPRNG) for security-critical code. This can be enabled by the option \texttt{secure\char`_mode}, which enables CSPRNG for noise generation and random batch composition. 

\textit{Noise scheduler and variable batch size.}
Similar to learning rate scheduler in deep learning, the noise scheduler in Opacus adjusts the noise multiplier during training by evolving it according to some predefined schedule, such as exponential, step, and custom function. Opacus also supports varying batch sizes throughout training.

\textit{Modular.}
Opacus integrates well with PyTorch Lightning, a high-level abstraction framework for PyTorch, which reduces boilerplate and simplifies coding complex networks.

\section{Benchmarks}
\label{sec:experiments}


We benchmark Opacus against other frameworks for training models with DP-SGD as well as standard PyTorch by comparing their respective runtimes on several end-to-end model training tasks.
We also quantify the runtime and memory overhead of training with DP using Opacus compared to training without DP using PyTorch for each layer that Opacus currently supports.

\subsection{End-to-end benchmarks}
Our end-to-end benchmarks are based on the Fast-DPSGD benchmarks~\cite{benchmark}. We evaluate Opacus on four end-to-end model training tasks against a JAX implementation of DP-SGD, a custom TensorFlow Privacy implementation, BackPACK, and PyVacy, as well as standard PyTorch without DP.

\subsubsection{Frameworks} 

JAX is a general-purpose framework for high-performance numerical computing that uses just-in-time (JIT) compilation and static graph optimization. We use the custom implementation of DP-SGD found in \cite{benchmark} and denote it as \textit{JAX (DP)}.

TensorFlow Privacy is a TensorFlow library for differentially private model training. We use the custom implementation with vectorization and XLA-driven JIT compilation, which outperforms both the custom TensorFlow Privacy implementation without XLA and the existing TensorFlow Privacy library in~\cite{benchmark}. We denote it as {\em Custom TFP (XLA)} for consistency with \cite{benchmark}. 

 To enable per-sample gradient extraction, BackPACK extends several PyTorch layers with support for efficient Jacobian computation. In contrast, PyVacy processes each sample individually in a for-loop, forgoing parallelization. 

\subsubsection{Experimental setup}

Following \cite{benchmark}, we train a CNN with 26,010 parameters on MNIST~\cite{mnist}, a handwritten digit recognition dataset, and a CNN with 605,226 parameters on CIFAR-10~\cite{cifar10}, a dataset of small color images. We train an embedding network and an LSTM network with 160,098 and 1,081,002 parameters respectively on the IMDb dataset~\cite{imdb}, which consists of movie reviews for sentiment classification. 
 
For each model and framework, we train the model using the framework's implementation of DP-SGD with a given privacy budget.
Since Opacus is built on top of PyTorch, we also train each model using PyTorch \textit{without} DP to better understand the runtime overhead of enabling DP with Opacus compared to training without DP using PyTorch.

We benchmark the latest version of each framework as of December 8th, 2021 in a separate Docker container, 
see~\cref{sec:exp-details} for details. Compared to the setup in~\cite{benchmark}, our GPU has more VRAM (40GB rather than 12GB), which allows us to benchmark larger batch sizes (512, 1024, 2048) for a more extensive comparison. The code is available at \url{https://github.com/TheSalon/fast-dpsgd/tree/latest}. 

We report each framework's median per-epoch runtime on each end-to-end training task at various batch sizes in \cref{fig:results}.

\begin{table}
\footnotesize
\centering
\begin{tabular}{lrrrrrrrr} 
	\toprule
	Batch Size & 16 & 32 & 64 & 128 & 256 & 512 & 1024 & 2048 \\ 
	\midrule
	JAX (DP) & 3.72 & 1.93 & 0.94 & 0.52 & 0.26 & 0.18 & 0.16 & 0.15\\
    \textit{PyTorch without DP} & \textit{5.82} & \textit{2.97} & \textit{1.55} & \textit{0.82} & \textit{0.47} & \textit{0.26} & \textit{0.16} & \textit{0.11}\\
    Opacus & 15.81 & 8.00 & 4.25 & 2.30 & 1.22 & 0.64 & 0.36 & 0.21\\
    BackPACK & 21.41 & 11.00 & 5.62 & 3.14 & 1.83 & 1.31 & 1.41 & 1.31\\
    Custom TFP (XLA) & 13.55 & 10.61 & 9.05 & 8.30 & 7.87 & 7.56 & 7.32 & 7.44\\
    PyVacy & 109.08 & 106.31 & 107.96 & 108.47 & 109.13 & 110.94 & 110.15 & 112.28\\
	\bottomrule
	\vspace{-0.05in}\\
	\multicolumn{9}{c}{(a) MNIST with CNN \vspace{0.15in}}\\

	\toprule
	Batch Size & 16 & 32 & 64 & 128 & 256 & 512 & 1024 & 2048 \\ 
	\midrule
    JAX (DP) & 10.20 & 5.98 & 3.64 & 3.28 & 2.91 & 2.75 & 2.66 & 2.60\\
    \textit{PyTorch without DP} & \textit{11.07} & \textit{5.99} & \textit{3.42} & \textit{1.76} & \textit{1.07} & \textit{0.87} & \textit{0.82} & \textit{0.79}\\
    Opacus & 32.02 & 16.59 & 8.66 & 4.40 & 2.45 & 2.06 & 1.93 & 1.89\\
    BackPACK & 46.57 & 24.17 & 13.09 & 10.66 & 10.91 & 10.36 & 10.06 & 10.02\\
    Custom TFP (XLA) & 41.51 & 37.45 & 33.66 & 33.07 & 32.18 & 30.78 & 31.82 & 30.73\\
    PyVacy & 198.35 & 196.63 & 192.54 & 191.75 & 192.27 & 195.36 & 192.90 & 192.76\\
	\bottomrule
	\vspace{-0.05in}\\
	\multicolumn{9}{c}{(b) CIFAR-10 with CNN \vspace{0.15in}}\\

	\toprule
	Batch Size & 16 & 32 & 64 & 128 & 256 & 512 & 1024 & 2048 \\ 
	\midrule
	JAX (DP) & 0.74 & 0.40 & 0.19 & 0.10 & 0.06 & 0.05 & 0.05 & 0.04\\
    \textit{PyTorch without DP} & \textit{1.54} & \textit{0.78} & \textit{0.40} & \textit{0.21} & \textit{0.11} & \textit{0.06} & \textit{0.04} & \textit{0.04}\\
    Opacus & 3.49 & 1.76 & 0.88 & 0.45 & 0.23 & 0.12 & 0.11 & 0.12\\
    Custom TFP (XLA) & 1.93 & 1.11 & 0.68 & 0.47 & 0.37 & 0.31 & 0.29 & 0.29\\
    PyVacy & 18.73 & 18.71 & 18.43 & 18.58 & 18.63 & 18.93 & 18.00 & 18.65\\
	\bottomrule
	\vspace{-0.05in}\\
	\multicolumn{9}{c}{(c) IMDb with Embedding \vspace{0.15in}}\\

	\toprule
	Batch Size & 16 & 32 & 64 & 128 & 256 & 512 & 1024 & 2048 \\ 
	\midrule
    JAX (DP) & 25.61 & 12.89 & 8.68 & 6.48 & 5.23 & 4.54 & 4.33 & 4.22\\
    \textit{PyTorch without DP} & \textit{14.08} & \textit{7.13} & \textit{3.91} & \textit{2.00} & \textit{1.04} & \textit{0.60} & \textit{0.41} & \textit{0.34}\\
    Opacus & 358.80 & 183.22 & 118.11 & 59.92 & 30.45 & 15.91 & 11.51 & 9.94\\
    Custom TFP (XLA) & 21.99 & 12.30 & 8.80 & 5.03 & 3.08 & 2.21 & 1.65 & 1.44\\
    PyVacy & 201.27 & 201.67 & 200.49 & 200.36 & 200.48 & 201.55 & 200.63 & 202.34\\
	\bottomrule
	\vspace{-0.05in}\\
	\multicolumn{9}{c}{(d) IMDb with LSTM \vspace{0.15in}}\\
\end{tabular}
\caption{Median runtime (in seconds) per epoch, computed over 20 epochs, of several DP-SGD frameworks as well as PyTorch without DP on four end-to-end model training tasks at various batch sizes. Since BackPACK does not support embedding or LSTM layers, the corresponding rows are omitted.}
\label{fig:results}
\end{table}


\subsubsection{Results}

JAX (DP) consistently achieves the lowest runtime among all DP-SGD frameworks on both MNIST and IMDb with the embedding network, even outperforming PyTorch without DP at smaller batch sizes. On CIFAR-10, JAX (DP) outperforms all other frameworks at smaller batch sizes, while Opacus surpasses JAX (DP) at larger batch sizes. On IMDb with the LSTM network, Custom TFP (XLA) consistently achieves the lowest runtime among all DP-SGD frameworks. Training with PyVacy consistently results in the highest runtime due to its use of micro-batching.

At a batch size of 2048, Opacus achieves the lowest runtime on CIFAR-10 and the second lowest runtime after JAX (DP) on MNIST and IMDb with the embedding network, its runtime being 1.4$\times$ and 3$\times$ that of JAX (DP) respectively. On IMDb with the LSTM network, Opacus achieves the third lowest runtime after Custom TFP (XLA) and JAX (DP), with 7$\times$ the runtime of Custom TFP (XLA) and 2.4$\times$ the runtime of JAX (DP).

While the median per-epoch runtime decreases as the batch size increases for most frameworks, the effect is strongest for Opacus and PyTorch: By increasing the batch size from 16 to 2048, Opacus's per-epoch runtime decreases by a factor ranging from 17$\times$ (on CIFAR-10) to 75$\times$ (on MNIST).
We report the mean per-epoch runtime reduction from increasing the batch size from 16 to 2048 
for each framework: 40$\times$ (Opacus), 37.8$\times$ (PyTorch without DP), 12.8$\times$ (JAX (DP)), 10.5$\times$ (BackPACK), 6.3$\times$ (Custom TFP (XLA)), and 1$\times$ (PyVacy). 

Since Opacus and PyTorch benefit the most from a larger batch size, increasing the batch size further may close the gap (where applicable) between Opacus and JAX (DP) or Custom TFP (XLA) and even result in Opacus outperforming them.
Both JAX (DP) and Custom TFP (XLA) rely on JIT compilation, which incurs a large runtime overhead in the first epoch (up to 101$\times$ and 625$\times$ the median per-epoch runtime, respectively) in exchange for a lower runtime in subsequent epochs.
See~\cref{fig:cumul} for each framework's cumulative runtime over 20 epochs.

On MNIST, CIFAR-10, and IMDb with the embedding network, enabling DP with Opacus incurs a 2$\times$ to 2.9$\times$ runtime overhead compared to training without DP using PyTorch. On IMDb with the LSTM network, the runtime overhead ranges from 25$\times$ to 30$\times$, see~\cref{sec:microresults} for further analysis. 
Since Opacus is built on PyTorch, we expect 
any future improvements to PyTorch's efficiency (e.g. \texttt{torch.vmap} graduating from the prototype stage) to benefit Opacus as well.

\subsection{Microbenchmarks}
We measure the runtime and peak allocated memory for one forward and one backward pass for each layer currently supported by Opacus, both with and without DP. We report the runtime and memory overhead of enabling DP with Opacus for each layer. 

\subsubsection{Experimental setup}
For each layer that Opacus currently supports, we benchmark both the layer with DP enabled and the corresponding \texttt{torch.nn} module without DP at various batch sizes. 

For the convolutional, normalization, linear, and embedding layers, which Opacus supports directly, wrapping the corresponding \texttt{torch.nn} module in Opacus's \texttt{GradSampleModule} enables DP. For the multi-head attention and RNN-based layers, which Opacus provides custom implementations for, wrapping the corresponding custom module in \texttt{GradSampleModule} enables DP.

Each benchmark estimates the mean runtime and peak allocated CUDA memory for one forward and one backward pass by measuring the cumulative runtime and peak allocated CUDA memory for a total of 2{,}000 forward and backward passes on 100 different input batches.
See~\cref{sec:exp-details} for details. The microbenchmarking code is available at \url{https://github.com/pytorch/opacus/tree/main/benchmarks}.

\begin{figure}[t]
     \centering
     \includegraphics[width=\linewidth]{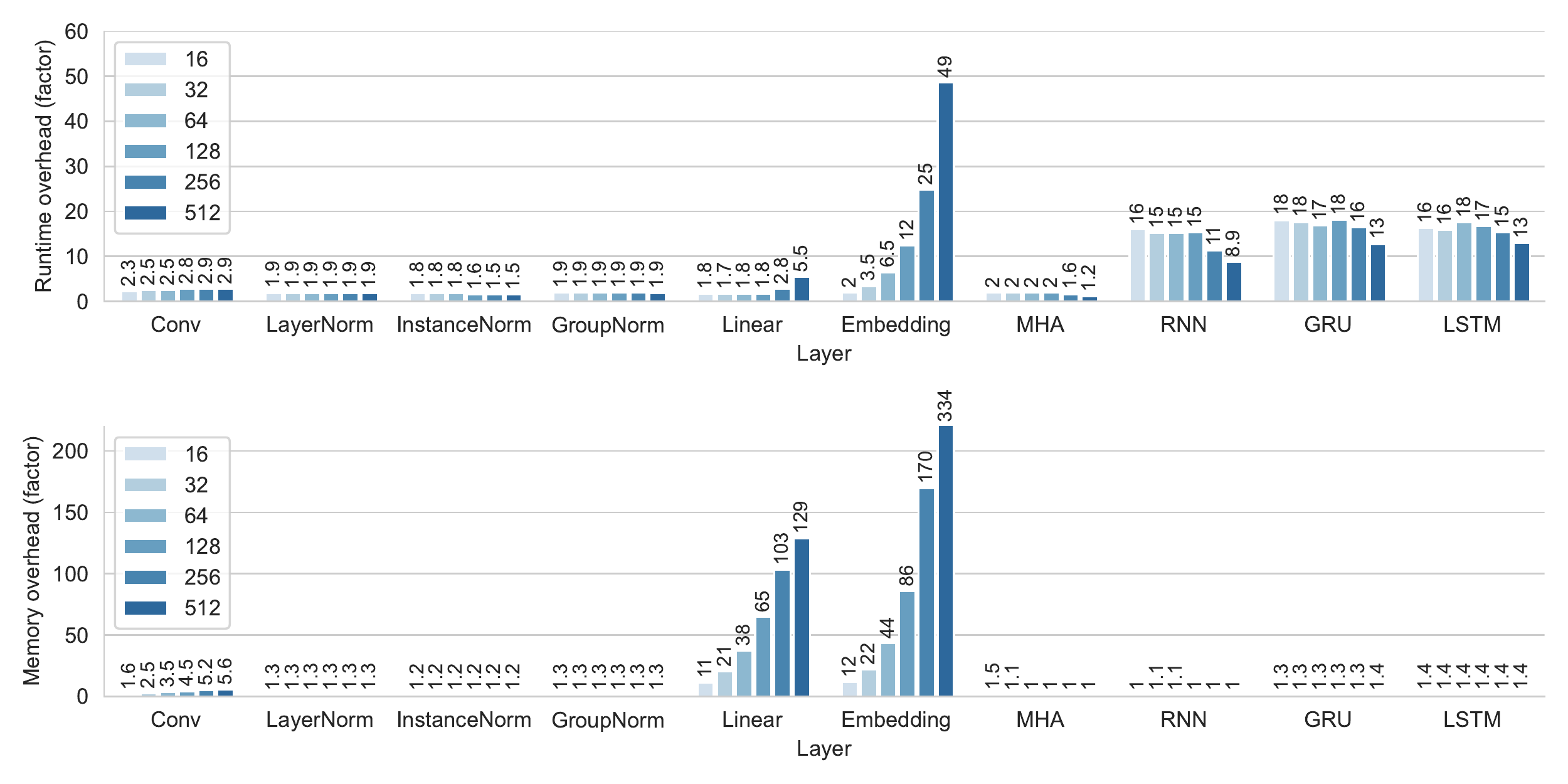}

    \caption{Runtime and peak allocated memory overhead of enabling DP for each layer currently supported by Opacus at various batch sizes. Top: Runtime overhead (factor). Bottom: Peak allocated memory overhead (factor). The runtime overhead is the mean runtime for one forward and one backward pass of the DP-enabled layer divided by the mean runtime for one forward and one backward pass of the corresponding \texttt{torch.nn} module without DP. The overhead in terms of peak allocated memory is calculated in the same manner.} 
    \label{fig:microbe}
\end{figure}

We report the runtime and peak memory overhead of the DP-enabled layer relative to the corresponding \texttt{torch.nn} module for all currently supported layers in \cref{fig:microbe}. 

\subsubsection{Memory requirements of DP}

Since DP-SGD requires per-sample gradients, training with DP-SGD uses more memory than training without DP-SGD. Let $L$ denote the number of trainable parameters in a given module 
and let each parameter be of size 1. Let $C$ be the size of the features, label, and model output for a single data point. Let $M$ denote the total memory usage for one forward and one backward pass on a batch of size $b$. Then, ignoring intermediate computations and constant additive overhead such as from non-trainable parameters:
\begin{align}
    M_\text{non-DP} &= bC + 2L,\\
    M_\text{DP} &=  bC + (1 + b)L.
    \label{eq:mem1}
\end{align}
In both non-DP and DP training, the features, labels, and the module's output for $b$ data points occupy memory of size $bC$, and the module itself occupies memory of size $L$ by definition. Without DP, we expect the gradient to occupy memory of size $L$ as well, whereas with DP, we expect the gradient to occupy memory of size $bL$ due to $b$ per-sample gradients. 
For $b \gg 1$, we can approximate the memory overhead as follows:\footnote{Since $L/C \sim b \leftrightarrow L \sim bC$.}
\begin{align}
    \frac{M_\text{DP}}{M_\text{non-DP}} & \approx
    \begin{cases}
        \frac{bC + (1+b)L}{bC} \approx 1+\frac{L}{C} & \text{if } L/C \ll b\\
        \frac{2+b}{3} \approx \frac{b}{3} & \text{if } L/C \approx b\\
        \frac{1+b}{2} \approx \frac{b}2 & \text{if } L/C \gg b
    \end{cases}
    \label{eq:memory}
\end{align}

\subsubsection{Results}
\label{sec:microresults}

\textbf{Runtime.}
For the convolutional, normalization, and multi-head attention layers, enabling DP with Opacus's \texttt{GradSampleModule} results in a 1.2$\times$ to 2.9$\times$ runtime increase, which we attribute to the calculation of per-sample gradients. For the linear and embedding layers, the runtime overhead increases with the batch size, reaching a factor of up to 5.5$\times$ and  49$\times$ respectively.

In contrast, enabling DP for RNN-based layers consistently incurs a large (up to 18$\times$) runtime overhead, which decreases as the batch size increases.
Opacus's custom RNN-based modules are responsible for most of this overhead,
their runtime being up to 11$\times$ the runtime of the corresponding \texttt{torch.nn} module. As with directly supported \texttt{torch.nn} modules, wrapping the custom modules in \texttt{GradSampleModule} results in a \textasciitilde2$\times$ slowdown. \cref{fig:microbe_appendix} compares the runtime of the \texttt{torch.nn} module, the corresponding custom module without DP, and the latter wrapped in \texttt{GradSampleModule} with DP enabled for the multi-head attention and RNN-based layers.

In practice, Opacus's custom LSTM with DP enabled performs competitively with other DP-SGD frameworks when training with large batch sizes. Recall that on the IMDb dataset, Opacus's median per-epoch runtime is only 2.4$\times$ the median per-epoch runtime of the custom JAX DP-SGD implementation while avoiding the latter's 101$\times$ JIT compilation overhead in the first epoch (see \cref{fig:results}, \cref{fig:cumul}).

\textbf{Peak allocated CUDA memory.} 
For the normalization, multi-head attention, and RNN-based layers, enabling DP with Opacus's \texttt{GradSampleModule} results in an up to 1.5$\times$ increase in peak allocated memory. In our experiments, $L/C$ is much smaller than the batch size for these layers, hence the relatively constant memory overhead.

For the linear and embedding layers, $L/C$ is substantial compared to the batch size. Hence, as predicted by \cref{eq:memory} and depicted in \cref{fig:microbe}, the peak allocated memory overhead increases with the batch size, reaching factors of up to 129$\times$ and 334$\times$ respectively.

\begin{figure}[t]
    \centering
    \includegraphics[width=\linewidth]{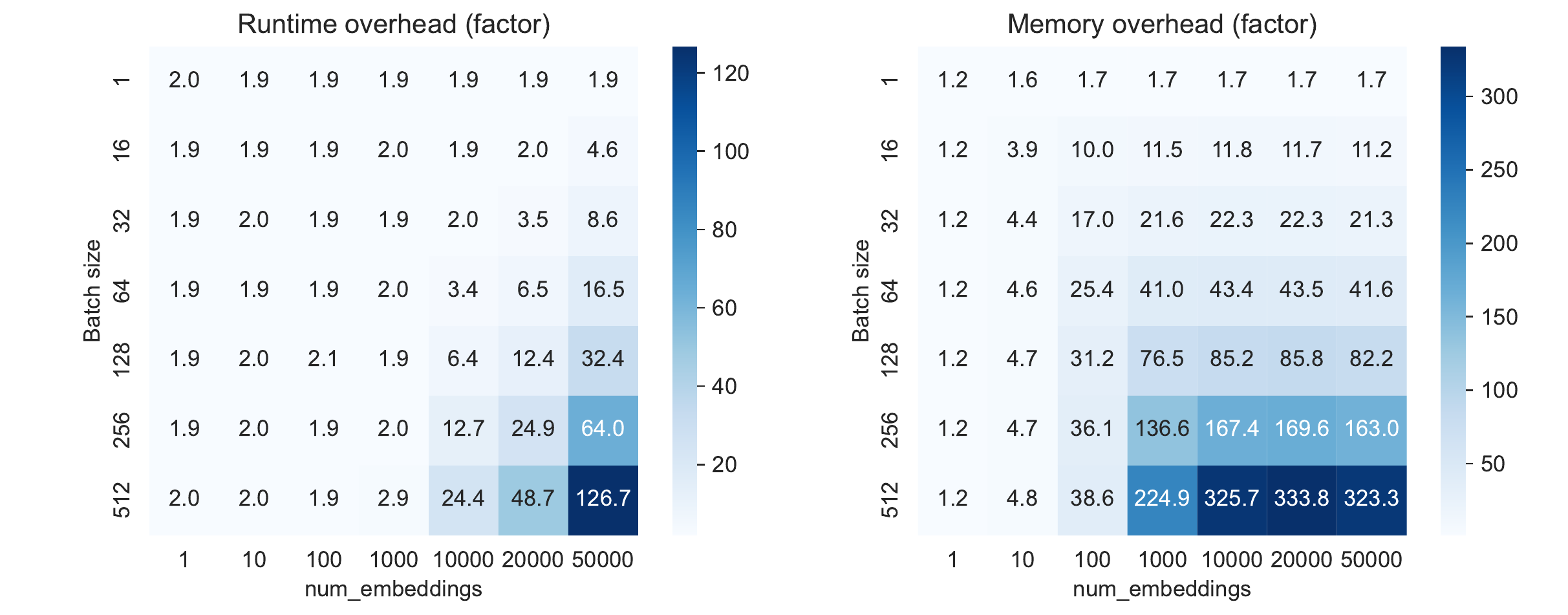}
    \caption{Runtime and peak allocated memory overhead of enabling DP for the embedding layer. 
    In addition to the batch size, we also vary \texttt{num\_embeddings} and thus, the size of the module $L$.
    Left: Runtime overhead (factor). Right: Peak allocated memory overhead (factor). For each value of \texttt{num\textunderscore embeddings}  from left to right, $L/C =  0.63, 5, 50, 496, 4{,}951, 9{,}901, 25{,}955$ respectively. Overheads are calculated as in \cref{fig:microbe}}
    \label{fig:embedding}
\end{figure}

\cref{fig:embedding} shows how $L/C$ and the batch size affect the runtime and peak allocated memory overhead of enabling DP by example of the embedding layer. Comparing the predicted memory overhead of enabling DP to the measured peak allocated memory overhead for various \texttt{num\_embeddings} and batch sizes shows that on average, \cref{eq:memory} overestimates the memory overhead of enabling DP by $41\% \pm 17.6\%$ when $L/C \ll b$ and underestimates it by $23.3\% \pm 6.3\%$ when $L/C \gg b$.


For the convolutional layer, the peak allocated memory overhead increases slightly with the batch size. As opposed to the linear and embedding layers, $L/C \ll b$ and the phenomenon weakens as the batch size increases. We attribute this to the comparatively many intermediate computations and additive overheads of the convolutional layer that are not captured in~\cref{eq:mem1}: For the other supported \texttt{torch.nn} layers, $M_\text{non-DP}$ and $M_\text{DP}$, on average, explain 56.5\% $\pm$ 8.7\% and 57.5\% $\pm$ 14.9\% of the measured peak allocated CUDA memory, whereas for the convolutional layer, $M_\text{non-DP}$ and $M_\text{DP}$ explain only 17.7\% $\pm$ 7.1\% and 6.15\% $\pm$ 0.03\% of the measured peak allocated memory.

\section{Related Work}
\label{sec:related}
\paragraph{Gradient Clipping.} At the heart of implementing DP-SGD is the need to compute clipped gradients, for which there are several different approaches. 
A first option, as implemented in Opacus, is to directly compute and clip the per-sample gradients.
A second option is to compute only the \textit{norm} of each sample's gradient (in an exact or approximate form), and form the weighted average loss over the batch by weighting samples according to their norm. 
Typically, each sample's weight is $C/\max(N_i, C)$, where $C$ is the clipping threshold and $N_i$ is the norm of the sample's gradient.
The gradient of this loss with respect to the parameters yields the average of clipped gradients. 
This option was proposed by Goodfellow~\cite{goodfellow} with exact norms, and was considered more recently along with Johnson-Lindenstrauss projections~\cite{bu2021fast} to compute approximate gradient norms.

Goodfellow's method is based on computing per-sample $\ell_2$-norms of the gradients and is restricted to fully-connected layers; more recently, Rochette et al.~\cite{rochette2019efficient} extended it to CNNs.  
Lee and Kifer~\cite{lee2021scaling} propose computing the {\em norm} of the per-sample gradients directly, hence doing two passes of back-propagation: one pass for obtaining the norm, and one pass for using the norm as a weight. 
In Opacus per-sample gradients are obtained in a single back-propagation pass, without performance or accuracy penalties of alternative techniques. 

Shortly after an initial version of this work was published, Li et al.~\cite{li2021large} generalized the Goodfellow method to handle sequential inputs, making fine-tuning large Transformers under DP computationally efficient. The proposed method, {\em ghost clipping}, is memory-efficient and has good throughput. In their experiments on large language models, Opacus is as good in memory efficiency and better in throughput than JAX, thanks to some improvements in the implementation. We plan to incorporate such improvements to optimize performance of Opacus further.

\paragraph{Frameworks for differentially private learning.}
TensorFlow Privacy and PyVacy are two existing frameworks providing implementation of DP-SGD for TensorFlow and PyTorch, respectively. 
BackPACK \cite{backpack}, another framework for DP-SGD, exploits Jacobians for efficiency. 
BackPACK currently supports only fully connected or convolutional layers, and several activation layers (recurrent and residual layers are not yet supported). 
Objax \cite{objax2020github} is a machine learning framework for JAX and also provides a DP-SGD implementation. The per-sample gradient clipping in Objax is based on a \texttt{vmap} method. It closely resembles the approach implemented in Subramani et al.~\cite{benchmark} and we believe benchmark findings are applicable to both implementations.
In~\cref{sec:experiments} we compare the performance of these frameworks with Opacus. 

Finally, we mention alternative approaches to ML training under different notions of privacy and security: 
using secure hardware to support oblivious training over encrypted data~\cite{Ohrimenko-oblivious}, or relying on secure multi-party computation techniques (SMPC) to train over data jointly held by several protocol participants~\cite{crypten}. 




\section{Conclusions}
\label{sec:conclusions}
Opacus is a PyTorch library for training deep learning models with differential privacy guarantees.  
The system design aims to provide simplicity, flexibility, and speed, for maximal compatibility with existing ML pipelines. 
We have outlined how these design principles have influenced the features of Opacus, and demonstrated that Opacus not only outperforms existing frameworks for training with differential privacy, but performs competitively with custom JIT compiled DP-SGD implementations on a variety of models and datasets.

Opacus is actively maintained as an open source project, supported primarily by the privacy-preserving machine learning team at Meta AI.  
A number of extensions and upgrades are planned for Opacus in the future, including enhanced flexibility for custom components, further efficiency improvements, and improved integration with the PyTorch ecosystem through projects like PyTorch Lightning.

\section*{Acknowledgements}
We are extremely grateful to Peter Romov for his substantial and valuable contributions to the Opacus codebase. Opacus also owes a debt of thanks to all of its open-source contributors who continuously helped in the development and maintaining of the Opacus library.

\printbibliography



\clearpage
\appendix
\section*{Appendix}

\section{Micro-Batching}
\label{sec:microbatching}
The following code snippet is a na\"ive way to yield the per-sample gradients through micro-batching.

{\small
\begin{verbatim}
for batch in Dataloader(train_dataset, batch_size):
    all_per_sample_gradients = [] 
    for x,y in batch:
        y_hat = model(x)
        loss = criterion(y_hat, y)
        loss.backward() 
        
        per_sample_grads = [p.grad.detach().clone() for p in model.parameters()]
        
        all_per_sample_gradients.append(per_sample_grads)
        model.zero_grad()  # reset p.grad
    
\end{verbatim}}

\section{Vectorized Computation}
\label{sec:vectorized}

In accordance with its speed objective, Opacus supports computing per-sample gradients efficiently, in a vectorized manner. 
This is achieved by deriving a per-sample gradient formula for every layer and transforming it into a form that can be implemented using a single application of the \texttt{einsum} operator. 
Due to space constraints, we discuss this approach only for the \texttt{nn.Linear} layer. 
The implementation details for other layers and other related tutorials can be found in \href{https://opacus.ai/tutorials}{\texttt{opacus.ai/tutorials}}.

Consider one linear layer with weight matrix $W$. 
We omit the bias from the forward pass equation and denote the forward pass by $Y=WX$, where $X$ is the input and $Y$ is the output of the linear layer. 
$X$ is a matrix of size $d\times B$, with $B$ columns ($B$ is the batch size), where each column is an input vector of dimension $d$. Similarly, the output matrix $Y$ would be of size $r\times B$ where each column is the output vector corresponding to an element in the batch and $r$ is the output dimension. 

The forward pass can be written as follows:
\[
Y_i^{(b)}=\sum_{j=1}^d W_{i,j} X_j^{(b)},
\]
where $Y_i^{(b)}$ denotes the $i$'th coordinate of the $b$'th sample in the batch.

In an ML problem, we typically need the derivative of the loss with respect to weights. Correspondingly, in Opacus we need the ``per-sample'' version of that, which is the per-sample derivative of the loss with respect to the weights $W$:
\[
  \frac{\partial L}{\partial z}=\sum_{b=1}^{B}\sum_{i'=1}^{r} \frac{\partial L}{\partial Y_{i'}^{(b)}} \frac{\partial Y_{i'}^{(b)}}{\partial z}.
\]
Applying the chain rule above, we can now replace variable $z$ with $W_{i,j}$ and get
\[
  \frac{\partial L}{\partial W_{i,j}}=\sum_{b=1}^{B}\sum_{i'=1}^{r} \frac{\partial L}{\partial Y_{i'}^{(b)}} \frac{\partial Y_{i'}^{(b)}}{\partial W_{i,j}} .
\]
We know from $Y=WX$ that $\frac{\partial Y_{i'}^{(b)}}{\partial W_{i,j}}$ is $X_j^{(b)}$ when $i=i'$, and is 0 otherwise. Continuing the above we have
\[
\frac{\partial L}{\partial W_{i,j}}=\sum_{b=1}^{B} \frac{\partial L}{\partial Y_{i'}^{(b)}} X_j^{(b)}.
\]

This equation corresponds to a matrix multiplication in PyTorch. In a regular backpropagation, the gradients of the loss function with respect to the weights (i.e., the gradients) are computed for the output of each layer and averaged over the batch. Since Opacus requires computing per-sample gradients, what we need is the following: 
\begin{equation}
\label{per-sample-eq}
\frac{\partial L_\mathit{batch}}{\partial W_{i,j}}=\frac{\partial L}{\partial Y_{i'}^{(b)}} X_j^{(b)}.
\end{equation}

More generally, in a neural network with more layers, equation (\ref{per-sample-eq}) can be written as 
\begin{equation}
\label{per-sample-eq-multi}
\frac{\partial L_\mathit{batch}}{\partial W^{(l)}_{i,j}}=\frac{\partial L}{\partial Z_{i}^{(l)(b)}} Z_j^{(l-1)(b)}
\end{equation}
for every layer $l$, where $Z_{i}^{(l)(b)}$ is the activation of the hidden layer $l$ for the $b$'th element of the batch of the neuron $i$. We refer to $\frac{\partial L}{\partial Z_{i}^{(l)(b)}}$ as the {\em highway gradient}. 

We now explain how we compute the per-sample gradient equation (\ref{per-sample-eq-multi}) in Opacus efficiently. In order to remove the sum reduction to get to the equations (\ref{per-sample-eq}) and (\ref{per-sample-eq-multi}), we need to replace the matrix multiplication with a batched outer product. In PyTorch, \texttt{einsum} allows us to do that in vectorized form. The function \texttt{einsum} computes multi-dimensional linear algebraic array operations by representing them in a short-hand format based on the Einstein summation convention. 

For instance, for computing the per-sample gradients for a linear layer, the \texttt{einsum} function can be written as \texttt{torch.einsum("n...i,n...j->nij", B, A)}, where variables \texttt{A} and \texttt{B} refer to activations and backpropagations, respectively. In Opacus activations and backpropagations essentially contain what we need for equation (\ref{per-sample-eq-multi}): using module and tensor hooks in PyTorch, Opacus stores the activations $Z_j^{(l-1)(b)}$ in forward hooks and access the highway gradients $\frac{\partial L}{\partial Z_{i}^{(l)(b)}}$ through backward hooks. That is how the method \texttt{torch.einsum("n...i,n...j->nij", B, A)} implements equation (\ref{per-sample-eq-multi}) for computing per-sample gradients for a \texttt{nn.Linear} layer. To understand the \texttt{einsum} expression itself, it is useful to think of it as a generalized version of \texttt{torch.bmm} (batch matrix multiplication) for multi-dimensional inputs. For 2D matrices \texttt{A} and \texttt{B}, \texttt{einsum} is equivalent to \texttt{torch.bmm(B.unsqueeze(2), A.unsqueeze(1))}. For higher dimensional inputs the idea is the same, while we also sum over extra dimensions. 

\section{Detection of DP Violations}
\label{sec:violation}

In this section we explain how Opacus can detect whether an operation violates some DP guarantees by applying the following criteria. First, Opacus checks if all layers in the model are supported. Second, Opacus checks for violations that make a model incompatible with differentially private training, which is usually due to one of the two issues: 1) the model tracks some extra information not covered by DP guarantees, or 2) a module is known to do batch-level computations, thus rendering the computation of per-sample gradients impossible. 

Some examples:
1) Opacus does not allow models to have batch normalization layers, as they share information across samples; 
2) Opacus does not allow \texttt{track\_running\_stats} in instance-normalization layers, as they track statistics that are not covered by DP guarantees.

The above checks in Opacus for DP compatibility are not exhaustive. In particular, Opacus has no way of checking whether the model maintains the independence of the individual samples or tracks extraneous statistics. We plan to investigate ways to address this in the future releases of Opacus.

\section{Tracking Gradients}
\label{sec:track-gradient}

In this section we explain how Opacus makes it easy to keep track of the gradient at different stages of DP training. In the following code snippet, we show how in Opacus we can access intermediate stages of gradient computation throughout training:

{\small
\begin{verbatim}
# model, optimizer and data_loader are initialized with make_private()

for data, labels in data_loader:
    output = model(data)
    loss = criterion(output, labels)
    loss.backward()
    
    print(p.grad) # normal gradients computed by PyTorch autograd
    print(p.grad_sample) # per-sample gradients computed by Opacus
                         # (no clipping, no noise)
    
    optimizer.step()

    print(p.grad_sample) # same as before optimizer.step() - this field is unchanged
    print(p.summed_grad) # clipped and aggregated over a batch, but no noise
    print(p.grad) # final gradients (clipped, noise added, aggregated over a batch)
    
    optimizer.zero_grad() # all gradients are None now
    
\end{verbatim}}

\section{Additional Experimental Results}
\subsection{End-to-end benchmarks}



\begin{figure}[h]
     \centering
     \begin{minipage}[t]{0.49\linewidth}
     \begin{minipage}[t]{0.99\linewidth}
         \includegraphics[width=\linewidth]{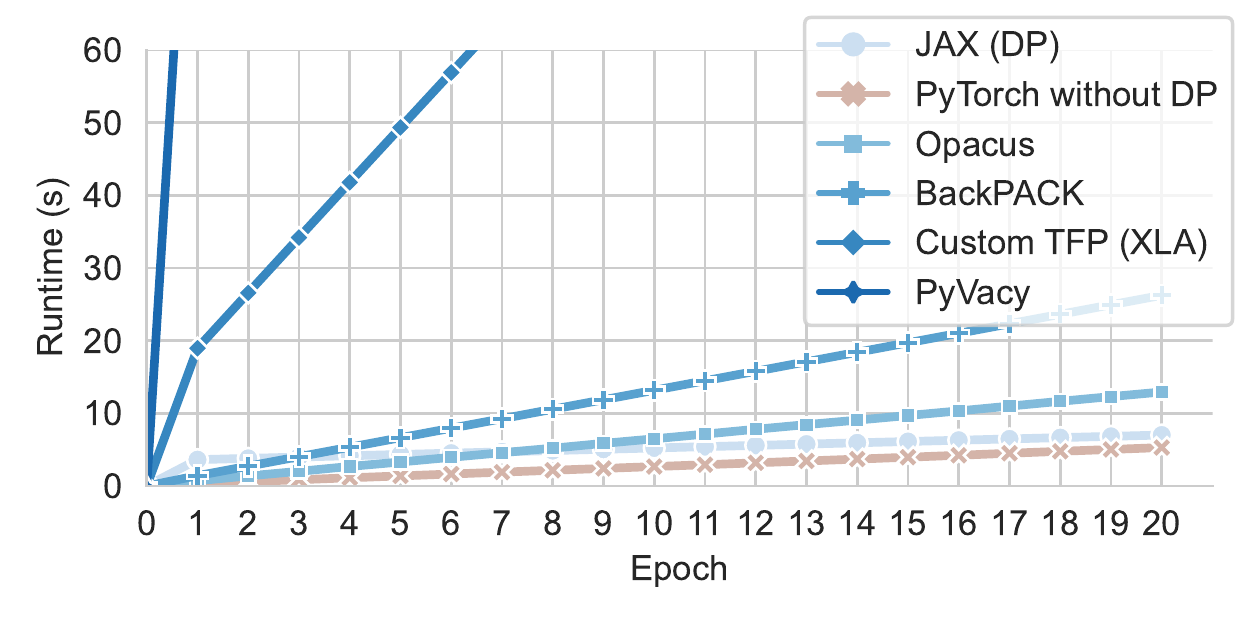}
         \caption*{\small (a) MNIST with CNN}
     \end{minipage}
     \begin{minipage}[t]{0.99\linewidth}
         \includegraphics[width=\linewidth]{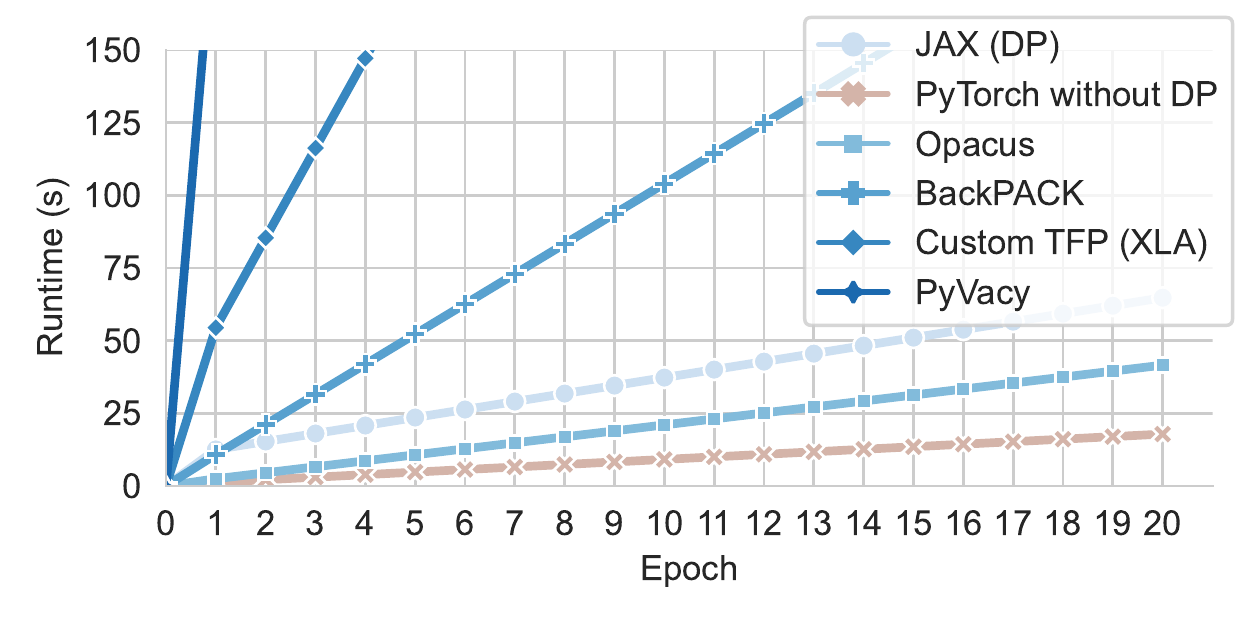}
         \caption*{\small (b) CIFAR-10 with CNN}
     \end{minipage}
      \end{minipage}
    \begin{minipage}[t]{0.49\linewidth}
    \begin{minipage}[t]{0.99\linewidth}
         \includegraphics[width=\linewidth]{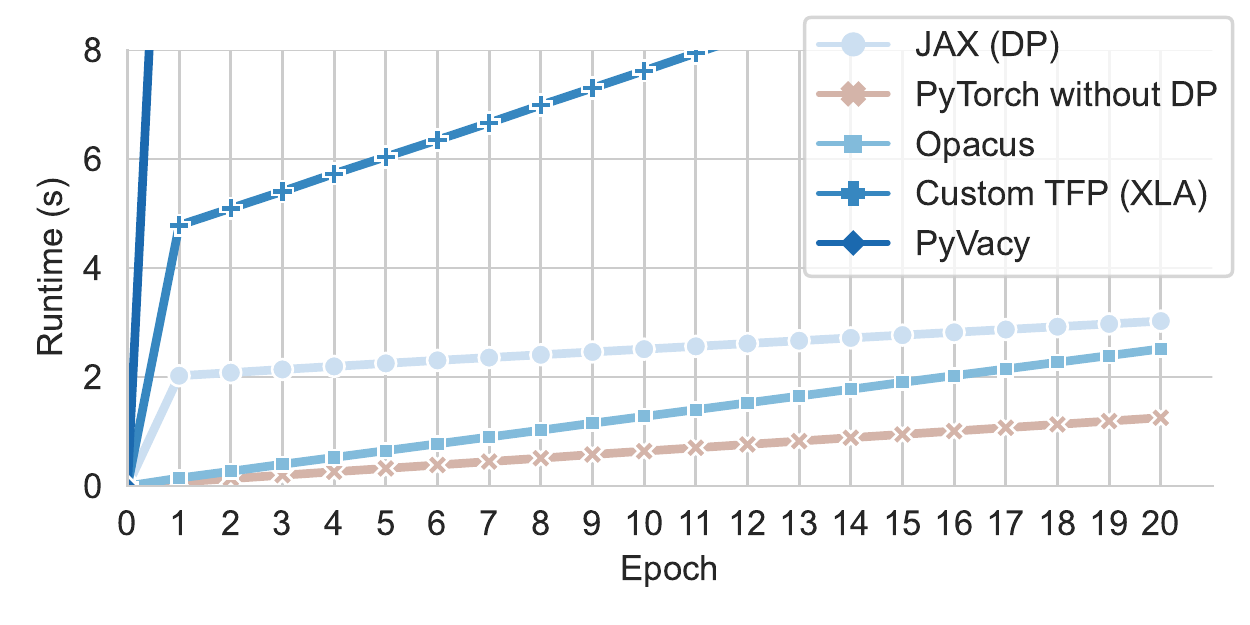}
         \caption*{\small (c) IMDb with Embedding}
     \end{minipage}
    \begin{minipage}[t]{0.99\linewidth}
         \includegraphics[width=\linewidth]{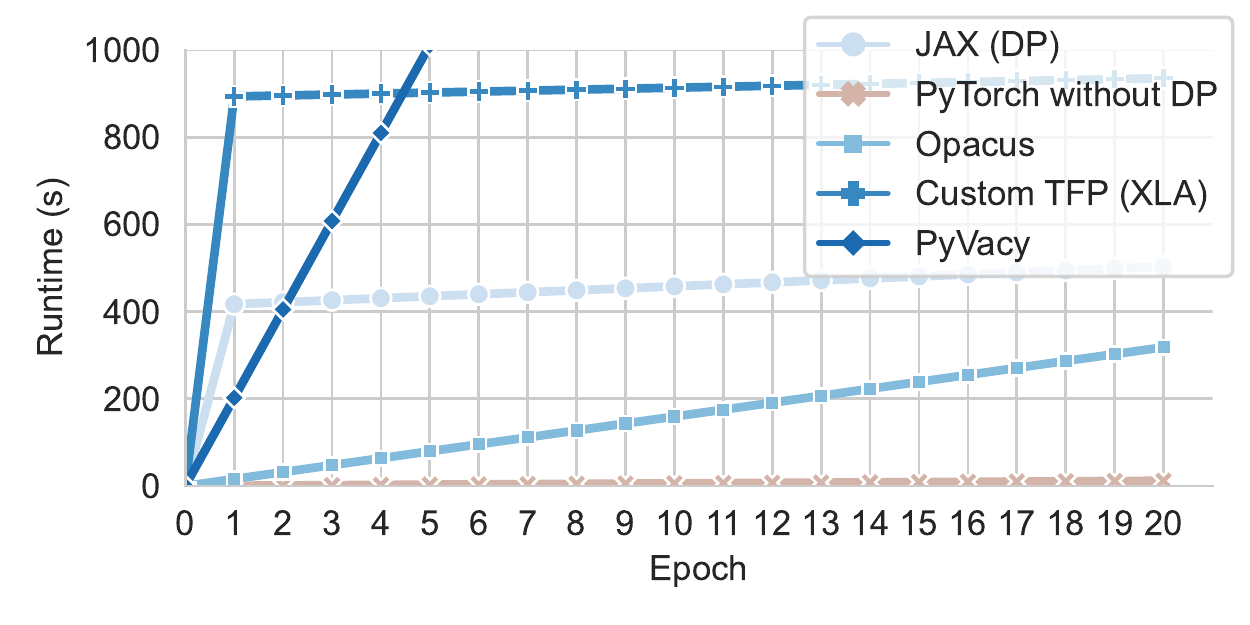}
         \caption*{\small (d) IMDb with LSTM}
     \end{minipage}
      \end{minipage}
\centering
\caption{Cumulative runtime over 20 epochs with batch size 512 for each framework. Using JIT compilation results in a slower first epoch.}
\label{fig:cumul}
\end{figure}


\cref{fig:cumul} shows each framework's cumulative runtime over 20 epochs with batch size 512 on each end-to-end model training task. Both JAX (DP) and Custom TFP (XLA) incur a large runtime overhead during the first epoch of up to 101$\times$ and 625$\times$ the runtime of subsequent epochs respectively due to JIT compilation. If training for relatively few epochs, disabling JIT compilation or using a framework that is optimized to run without JIT may reduce total runtime.

\subsection{Microbenchmarks}

\begin{figure}[t]
     \centering
     \begin{minipage}[t]{0.49\linewidth}
     \begin{minipage}[t]{0.99\linewidth}
         \includegraphics[width=\linewidth]{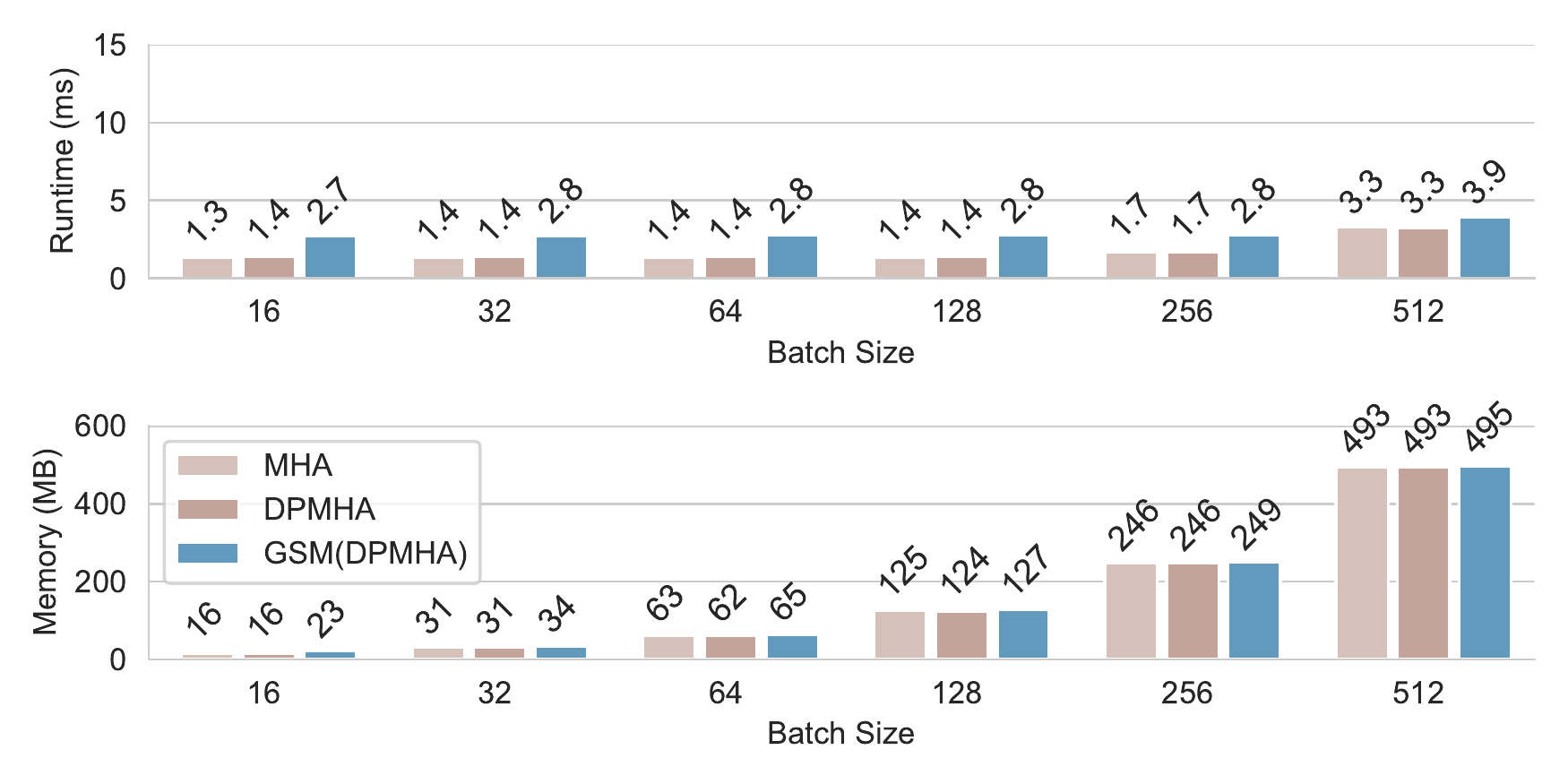}
         \caption*{\small (a) Multi-head Attention}
     \end{minipage}
     \begin{minipage}[t]{0.99\linewidth}
         \includegraphics[width=\linewidth]{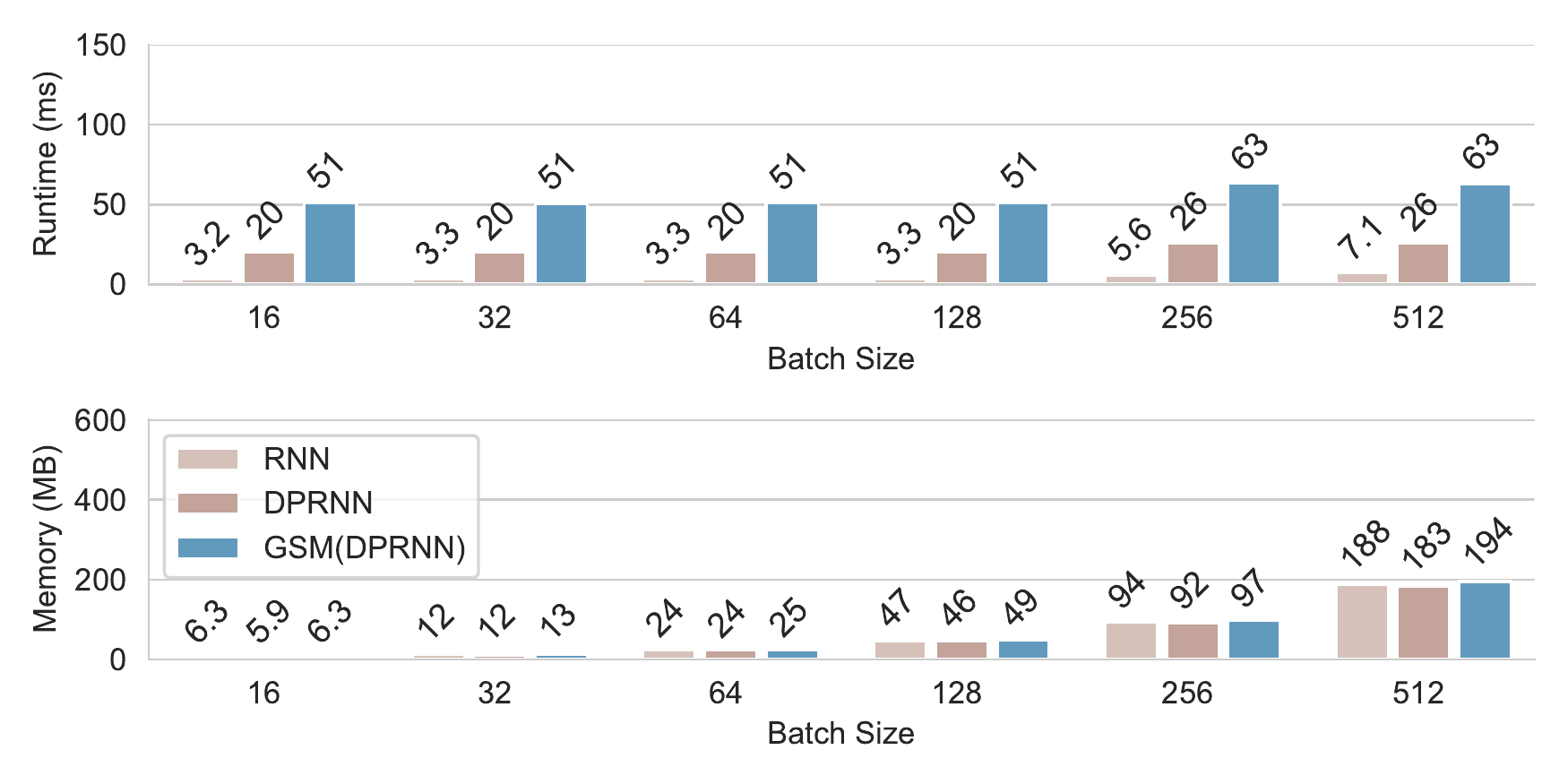}
         \caption*{\small (b) RNN}
     \end{minipage}
      \end{minipage}
    \begin{minipage}[t]{0.49\linewidth}
    \begin{minipage}[t]{0.99\linewidth}
         \includegraphics[width=\linewidth]{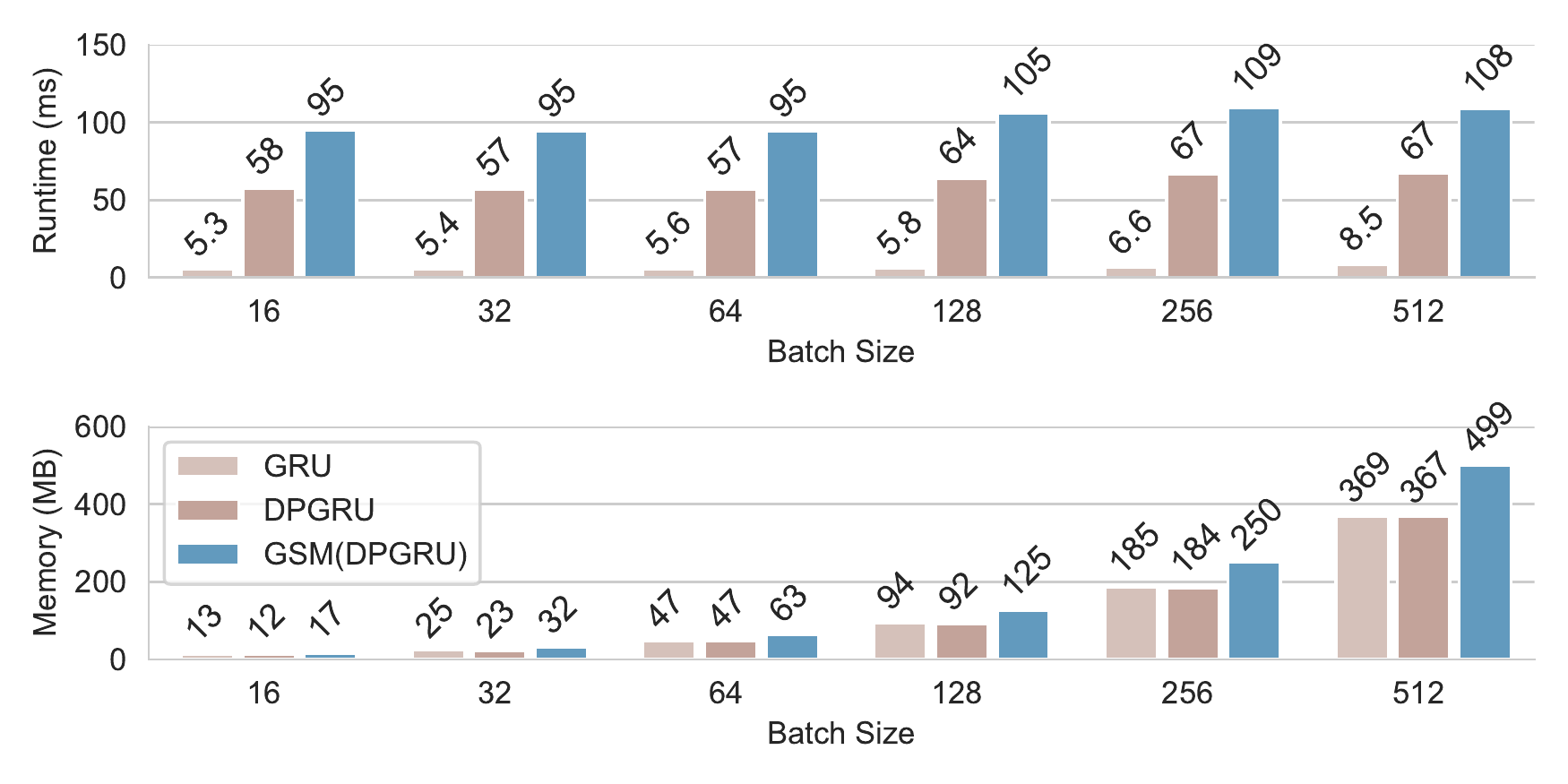}
         \caption*{\small (c) GRU}
     \end{minipage}
    \begin{minipage}[t]{0.99\linewidth}
         \includegraphics[width=\linewidth]{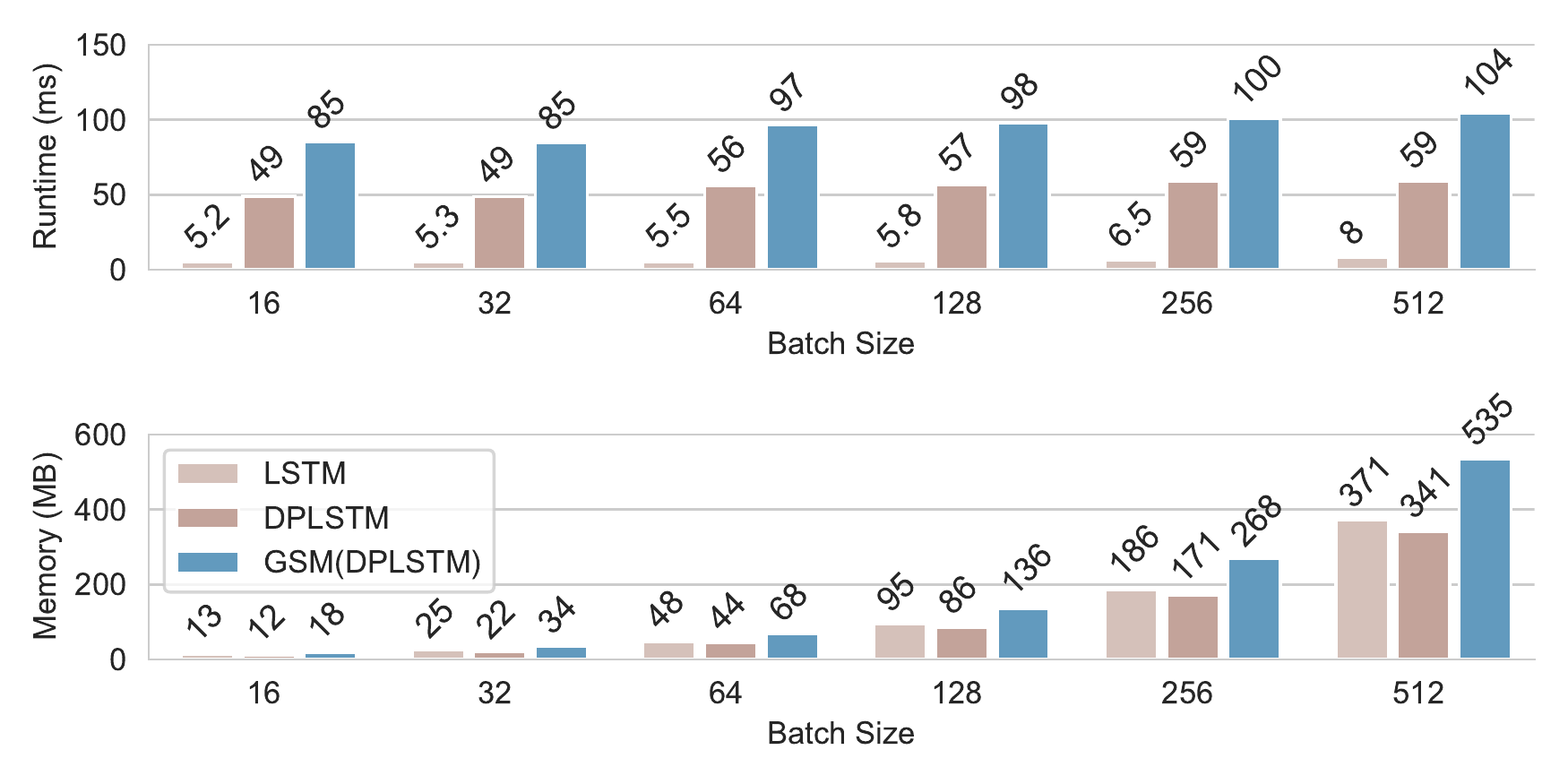}
         \caption*{\small (d) LSTM}
     \end{minipage}
      \end{minipage}
    
    \caption{Comparing the \texttt{torch.nn} module, the corresponding Opacus module without DP, and the Opacus module wrapped in \texttt{GradSampleModule} with DP enabled for the multi-head attention, RNN, GRU, and LSTM layers. Top: Mean runtime (ms). Bottom: Peak allocated memory (MB).
    }
    \label{fig:microbe_appendix}
\end{figure}

Opacus provides custom implementations for the multi-head attention, RNN, GRU, and LSTM layers, which can be wrapped in \texttt{GradSampleModule} to enable training with DP. \cref{fig:microbe_appendix} compares the runtime and peak memory usage of the \texttt{torch.nn} module, the corresponding Opacus module without DP, and the corresponding Opacus module wrapped in \texttt{GradSampleModule} with DP enabled for these layers.

For RNN-based layers, Opacus's custom modules are responsible for most of the runtime overhead of enabling DP, as they are up to 11$\times$ slower than the corresponding \texttt{torch.nn} module. Wrapping the custom modules in \texttt{GradSampleModule} results in a \textasciitilde2$\times$ slowdown.

The small peak allocated memory overhead (where applicable) is purely due to wrapping the custom modules in \texttt{GradSampleModule} and collecting per-sample gradients, as Opacus's custom modules tend to use slightly less memory than the corresponding \texttt{torch.nn} modules. 

\cref{tab:runtime} (runtime) and \cref{tab:memory} (memory) include the raw data used to generate \cref{fig:microbe} and \cref{fig:microbe_appendix}. \cref{tab:memory_breakdown} includes a breakdown of CUDA memory usage, as well as $L/C$ and $(L/C) / b$ for each layer and batch size. 

\begin{table}
\footnotesize
\begin{center}
\vspace{-0.7in}
\begin{tabular}{l| r| r| r| r| r} 
    \toprule
    & & \multicolumn{3}{c|}{\textbf{Runtime (ms)}} &\\
     \textbf{Layer} & \textbf{Batch size} & \texttt{nn.module} & \texttt{DPModule} & \texttt{GSM(module)} & \textbf{Factor (DP)}\\
    \midrule
    \multirow{6}{*}{Conv} & 16 & 2.66 & \multirow{6}{*}{n/a} & 6.18 & 2.32\\
     & 32 & 4.38 &  & 10.9 & 2.50\\
     & 64 & 8.71 &  & 21.7 & 2.49\\
     & 128 & 14.3 &  & 40.0 & 2.79\\
     & 256 & 27.6 &  & 79.0 & 2.87\\
     & 512 & 54.8 &  & 157 & 2.87\\
    \midrule
    \multirow{6}{*}{LayerNorm} & 16 & 0.27 & \multirow{6}{*}{n/a} & 0.52 & 1.92\\
     & 32 & 0.27 &  & 0.52 & 1.91\\
     & 64 & 0.27 &  & 0.52 & 1.91\\
     & 128 & 0.27 &  & 0.52 & 1.91\\
     & 256 & 0.27 &  & 0.52 & 1.90\\
     & 512 & 0.27 &  & 0.52 & 1.91\\
    \midrule
    \multirow{6}{*}{InstanceNorm} & 16 & 0.41 & \multirow{6}{*}{n/a} & 0.74 & 1.81\\
     & 32 & 0.41 &  & 0.74 & 1.81\\
     & 64 & 0.41 &  & 0.75 & 1.81\\
     & 128 & 0.73 &  & 1.13 & 1.55\\
     & 256 & 1.43 &  & 2.20 & 1.54\\
     & 512 & 2.77 &  & 4.28 & 1.54\\
    \midrule
    \multirow{6}{*}{GroupNorm} & 16 & 0.30 & \multirow{6}{*}{n/a} & 0.59 & 1.94\\
     & 32 & 0.30 &  & 0.59 & 1.94\\
     & 64 & 0.30 &  & 0.59 & 1.94\\
     & 128 & 0.30 &  & 0.59 & 1.95\\
     & 256 & 0.30 &  & 0.59 & 1.95\\
     & 512 & 0.31 &  & 0.60 & 1.91\\
    \midrule
    \multirow{6}{*}{Linear} & 16 & 0.37 & \multirow{6}{*}{n/a} & 0.65 & 1.76\\
     & 32 & 0.38 &  & 0.66 & 1.75\\
     & 64 & 0.38 &  & 0.66 & 1.76\\
     & 128 & 0.38 &  & 0.68 & 1.79\\
     & 256 & 0.39 &  & 1.08 & 2.80\\
     & 512 & 0.37 &  & 2.05 & 5.49\\
    \midrule
    \multirow{6}{*}{Embedding} & 16 & 0.25 & \multirow{6}{*}{n/a} & 0.50 & 2.01\\
     & 32 & 0.25 &  & 0.86 & 3.45\\
     & 64 & 0.25 &  & 1.62 & 6.52\\
     & 128 & 0.25 &  & 3.13 & 12.4\\
     & 256 & 0.25 &  & 6.15 & 24.9\\
     & 512 & 0.25 &  & 12.2 & 48.7\\
    \midrule
    \multirow{6}{*}{MHA} & 16 & 1.34 & 1.41 & 2.72 & 2.02\\
     & 32 & 1.36 & 1.43 & 2.76 & 2.02\\
     & 64 & 1.37 & 1.43 & 2.76 & 2.02\\
     & 128 & 1.36 & 1.43 & 2.77 & 2.04\\
     & 256 & 1.70 & 1.72 & 2.78 & 1.64\\
     & 512 & 3.29 & 3.27 & 3.91 & 1.19\\
    \midrule
    \multirow{6}{*}{RNN} & 16 & 3.20 & 20.4 & 51.3 & 16.0\\
     & 32 & 3.32 & 20.0 & 50.7 & 15.3\\
     & 64 & 3.34 & 20.0 & 50.8 & 15.2\\
     & 128 & 3.33 & 20.0 & 51.1 & 15.4\\
     & 256 & 5.59 & 25.8 & 63.1 & 11.3\\
     & 512 & 7.08 & 25.6 & 62.8 & 8.87\\
    \midrule
    \multirow{6}{*}{GRU} & 16 & 5.25 & 57.5 & 95.0 & 18.1\\
     & 32 & 5.36 & 56.8 & 94.6 & 17.7\\
     & 64 & 5.61 & 56.9 & 94.7 & 16.9\\
     & 128 & 5.84 & 63.9 & 106 & 18.2\\
     & 256 & 6.64 & 66.8 & 110 & 16.5\\
     & 512 & 8.51 & 67.2 & 109 & 12.8\\
    \midrule
    \multirow{6}{*}{LSTM} & 16 & 5.21 & 49.0 & 85.1 & 16.3\\
     & 32 & 5.32 & 48.8 & 84.8 & 15.9\\
     & 64 & 5.48 & 56.0 & 96.5 & 17.6\\
     & 128 & 5.84 & 56.6 & 97.8 & 16.7\\
     & 256 & 6.55 & 59.0 & 101 & 15.4\\
     & 512 & 8.04 & 59.1 & 105 & 13.0\\
     \bottomrule
\end{tabular}
\end{center}
\caption{Mean runtime (in milliseconds) for one forward and one backward pass for the \texttt{torch.nn} module, the corresponding Opacus module without DP (where applicable), and the corresponding Opacus or \texttt{torch.nn} module wrapped in \texttt{GradSampleModule} with DP enabled. The factor is the runtime of the layer with DP enabled divided by the runtime of the \texttt{torch.nn} module.}
\label{tab:runtime}
\end{table}
\begin{table}
\footnotesize
\begin{center}
\vspace{-0.75in}
\begin{tabular}{l| r| r| r| r| r} 
    \toprule
    & & \multicolumn{3}{c|}{\textbf{Peak allocated CUDA memory (MB)}} &\\
     \textbf{Layer} & \textbf{Batch size} & \texttt{nn.module} & \texttt{DPModule} & \texttt{GSM(module)} & \textbf{Factor (DP)}\\
     \midrule
    \multirow{6}{*}{Conv} & 16 & 738 & \multirow{6}{*}{n/a} & 1157 & 1.57\\
     & 32 & 929 &  & 2312 & 2.49\\
     & 64 & 1307 &  & 4621 & 3.54\\
     & 128 & 2065 &  & 9240 & 4.48\\
     & 256 & 3582 &  & 18479 & 5.16\\
     & 512 & 6615 &  & 36955 & 5.59\\
    \midrule
    \multirow{6}{*}{LayerNorm} & 16 & 0.03 & \multirow{6}{*}{n/a} & 0.04 & 1.32\\
     & 32 & 0.05 &  & 0.07 & 1.33\\
     & 64 & 0.10 &  & 0.14 & 1.33\\
     & 128 & 0.20 &  & 0.27 & 1.33\\
     & 256 & 0.40 &  & 0.53 & 1.33\\
     & 512 & 0.79 &  & 1.06 & 1.33\\
    \midrule
    \multirow{6}{*}{InstanceNorm} & 16 & 6.35 & \multirow{6}{*}{n/a} & 7.39 & 1.17\\
     & 32 & 12.7 &  & 14.8 & 1.17\\
     & 64 & 25.4 &  & 29.6 & 1.17\\
     & 128 & 50.7 &  & 59.1 & 1.17\\
     & 256 & 101 &  & 118 & 1.17\\
     & 512 & 203 &  & 236 & 1.17\\
    \midrule
    \multirow{6}{*}{GroupNorm} & 16 & 0.11 & \multirow{6}{*}{n/a} & 0.14 & 1.30\\
     & 32 & 0.21 &  & 0.27 & 1.32\\
     & 64 & 0.41 &  & 0.54 & 1.32\\
     & 128 & 0.81 &  & 1.07 & 1.32\\
     & 256 & 1.61 &  & 2.14 & 1.33\\
     & 512 & 3.22 &  & 4.27 & 1.33\\
    \midrule
    \multirow{6}{*}{Linear} & 16 & 3.28 & \multirow{6}{*}{n/a} & 36.9 & 11.2\\
     & 32 & 3.42 &  & 70.7 & 20.7\\
     & 64 & 3.68 &  & 138 & 37.6\\
     & 128 & 4.20 &  & 273 & 65.0\\
     & 256 & 5.25 &  & 543 & 103\\
     & 512 & 8.39 &  & 1083 & 129\\
    \midrule
    \multirow{6}{*}{Embedding} & 16 & 24.0 & \multirow{6}{*}{n/a} & 280 & 11.7\\
     & 32 & 24.0 & & 536 & 22.3\\
     & 64 & 24.1 & & 1048 & 43.5\\
     & 128 & 24.2 & & 2072 & 85.8\\
     & 256 & 24.3 & & 4122 & 170\\
     & 512 & 24.6 & & 8217 & 334\\
    \midrule
    \multirow{6}{*}{MHA} & 16 & 15.7 & 15.7 & 23.3 & 1.48\\
     & 32 & 31.3 & 31.3 & 33.9 & 1.08\\
     & 64 & 62.5 & 62.3 & 64.9 & 1.04\\
     & 128 & 125 & 125 & 127 & 1.02\\
     & 256 & 247 & 247 & 249 & 1.01\\
     & 512 & 493 & 493 & 496 & 1.01\\
    \midrule
    \multirow{6}{*}{RNN} & 16 & 6.27 & 5.94 & 6.35 & 1.01\\
     & 32 & 12.1 & 11.6 & 13.1 & 1.08\\
     & 64 & 23.9 & 23.5 & 25.1 & 1.05\\
     & 128 & 47.3 & 46.1 & 48.5 & 1.03\\
     & 256 & 94.2 & 92.0 & 97.3 & 1.03\\
     & 512 & 188 & 184 & 195 & 1.04\\
    \midrule
    \multirow{6}{*}{GRU} & 16 & 12.7 & 12.2 & 16.6 & 1.31\\
     & 32 & 24.6 & 23.4 & 32.0 & 1.30\\
     & 64 & 47.2 & 46.8 & 63.1 & 1.34\\
     & 128 & 94.2 & 92.3 & 125 & 1.33\\
     & 256 & 185 & 184 & 250 & 1.35\\
     & 512 & 369 & 368 & 499 & 1.35\\
    \midrule
    \multirow{6}{*}{LSTM} & 16 & 13.2 & 11.5 & 18.0 & 1.37\\
     & 32 & 24.7 & 22.0 & 34.5 & 1.40\\
     & 64 & 48.2 & 43.7 & 68.3 & 1.42\\
     & 128 & 94.8 & 85.9 & 136 & 1.44\\
     & 256 & 186 & 171 & 268 & 1.44\\
     & 512 & 371 & 342 & 536 & 1.44\\
     \bottomrule
\end{tabular}
\end{center}
\caption{Peak allocated CUDA memory (in MB) for one forward and one backward pass for the \texttt{torch.nn} module, the corresponding Opacus module without DP (where applicable), and the corresponding Opacus or \texttt{torch.nn} module wrapped in \texttt{GradSampleModule} with DP enabled. The factor is the peak allocated memory for the layer with DP enabled divided by the peak allocated memory for the \texttt{torch.nn} module.}
\label{tab:memory}
\end{table}
\begin{table}
\footnotesize
\begin{center}
\vspace{-0.7in}
\begin{tabular}{l| r| r| r| r| r| r| r}
    \toprule
     \textbf{Layer} & \textbf{Batch size} & \textbf{Input (MB)} & \textbf{Labels (MB)} & \textbf{$C$}& \textbf{Module ($L$)} & $L/C$ & $(L/C) / b$\\
    \midrule
    \multirow{6}{*}{Conv} & 16 & 21.0 & 16.4 & 3.36 & \multirow{6}{*}{1.05} & 0.31 & 0.02\\
     & 32 & 41.0 & 33.6 & 3.38 &  & 0.31 & 0.010\\
     & 64 & 81.9 & 65.5 & 3.33 &  & 0.32 & 0.005\\
     & 128 & 164 & 131 & 3.33 &  & 0.32 & 0.002\\
     & 256 & 328 & 262 & 3.33 &  & 0.32 & 0.001\\
     & 512 & 655 & 524 & 3.33 &  & 0.32 & 0.001\\
    \midrule
    \multirow{6}{*}{LayerNorm} & 16 & 0.004 & 0.004 & 0.001 & \multirow{6}{*}{0.001} & 1.33 & 0.08\\
     & 32 & 0.008 & 0.008 & 0.001 &  & 1.33 & 0.04\\
     & 64 & 0.02 & 0.02 & 0.001 &  & 1.33 & 0.02\\
     & 128 & 0.03 & 0.03 & 0.001 &  & 1.33 & 0.01\\
     & 256 & 0.07 & 0.07 & 0.001 &  & 1.33 & 0.005\\
     & 512 & 0.13 & 0.13 & 0.001 &  & 1.33 & 0.003\\
    \midrule
    \multirow{6}{*}{InstanceNorm} & 16 & 1.05 & 1.05 & 0.20 & \multirow{6}{*}{0.002} & 0.01 & 0.001\\
     & 32 & 2.10 & 2.10 & 0.20 &  & 0.01 & 0.000\\
     & 64 & 4.19 & 4.19 & 0.20 &  & 0.01 & 0.000\\
     & 128 & 8.39 & 8.39 & 0.20 &  & 0.01 & 0.000\\
     & 256 & 16.8 & 16.8 & 0.20 &  & 0.01 & 0.000\\
     & 512 & 33.6 & 33.6 & 0.20 &  & 0.01 & 0.000\\
    \midrule
    \multirow{6}{*}{GroupNorm} & 16 & 0.02 & 0.02 & 0.003 & \multirow{6}{*}{0.002} & 0.67 & 0.04\\
     & 32 & 0.03 & 0.03 & 0.003 &  & 0.67 & 0.02\\
     & 64 & 0.07 & 0.07 & 0.003 &  & 0.67 & 0.01\\
     & 128 & 0.13 & 0.13 & 0.003 &  & 0.67 & 0.005\\
     & 256 & 0.26 & 0.26 & 0.003 &  & 0.67 & 0.003\\
     & 512 & 0.52 & 0.52 & 0.003 &  & 0.67 & 0.001\\
    \midrule
    \multirow{6}{*}{Linear} & 16 & 0.03 & 0.03 & 0.006 & \multirow{6}{*}{1.05} & 171 & 10.7\\
     & 32 & 0.07 & 0.07 & 0.006 &  & 171 & 5.34\\
     & 64 & 0.13 & 0.13 & 0.006 &  & 171 & 2.67\\
     & 128 & 0.26 & 0.26 & 0.006 &  & 171 & 1.34\\
     & 256 & 0.52 & 0.52 & 0.006 &  & 171 & 0.67\\
     & 512 & 1.05 & 1.05 & 0.006 &  & 171 & 0.33\\
    \midrule
    \multirow{6}{*}{Embedding} & 16 & 0.001 & 0.007 & 0.001 & \multirow{6}{*}{8} & 9259 & 579\\
     & 32 & 0.001 & 0.01 & 0.001 &  & 9804 & 306\\
     & 64 & 0.001 & 0.03 & 0.001 &  & 9901 & 155\\
     & 128 & 0.001 & 0.05 & 0.001 &  & 9901 & 77.4\\
     & 256 & 0.002 & 0.10 & 0.001 &  & 9901 & 38.7\\
     & 512 & 0.004 & 0.20 & 0.001 &  & 9901 & 19.3\\
    \midrule
    \multirow{6}{*}{MHA} & 16 & 0.41 & 0.41 & 0.08 & \multirow{6}{*}{0.16} & 2.12 & 0.13\\
     & 32 & 0.82 & 0.82 & 0.08 &  & 2.12 & 0.07\\
     & 64 & 1.64 & 1.64 & 0.08 &  & 2.12 & 0.03\\
     & 128 & 3.28 & 3.28 & 0.08 &  & 2.12 & 0.02\\
     & 256 & 6.55 & 6.55 & 0.08 &  & 2.12 & 0.008\\
     & 512 & 13.1 & 13.1 & 0.08 &  & 2.12 & 0.004\\
    \midrule
    \multirow{6}{*}{RNN} & 16 & 0.82 & 0.82 & 0.15 & \multirow{6}{*}{0.08} & 0.53 & 0.03\\
     & 32 & 1.64 & 1.64 & 0.15 &  & 0.53 & 0.02\\
     & 64 & 3.28 & 3.28 & 0.15 &  & 0.53 & 0.008\\
     & 128 & 6.55 & 6.55 & 0.15 &  & 0.53 & 0.004\\
     & 256 & 13.1 & 13.1 & 0.15 &  & 0.53 & 0.002\\
     & 512 & 26.2 & 26.2 & 0.15 &  & 0.53 & 0.001\\
    \midrule
    \multirow{6}{*}{GRU} & 16 & 0.82 & 0.82 & 0.15 & \multirow{6}{*}{0.24} & 1.58 & 0.10\\
     & 32 & 1.64 & 1.64 & 0.15 &  & 1.58 & 0.05\\
     & 64 & 3.28 & 3.28 & 0.15 &  & 1.58 & 0.02\\
     & 128 & 6.55 & 6.55 & 0.15 &  & 1.58 & 0.01\\
     & 256 & 13.1 & 13.1 & 0.15 &  & 1.58 & 0.006\\
     & 512 & 26.2 & 26.2 & 0.15 &  & 1.58 & 0.003\\
    \midrule
    \multirow{6}{*}{LSTM} & 16 & 0.82 & 0.82 & 0.15 & \multirow{6}{*}{0.32} & 2.11 & 0.13\\
     & 32 & 1.64 & 1.64 & 0.15 &  & 2.11 & 0.07\\
     & 64 & 3.28 & 3.28 & 0.15 &  & 2.11 & 0.03\\
     & 128 & 6.55 & 6.55 & 0.15 &  & 2.11 & 0.02\\
     & 256 & 13.1 & 13.1 & 0.15 &  & 2.11 & 0.008\\
     & 512 & 26.2 & 26.2 & 0.15 &  & 2.11 & 0.004\\
     \bottomrule
\end{tabular}
\end{center}
\caption{Measured sizes (in MB) for the input, labels/output, and the \texttt{torch.nn} module for each layer and batch size. These numbers are (almost) identical for the corresponding Opacus module ($\pm 1\%$) and when wrapped in \texttt{GradSampleModule}. $C$ is calculated as the size of the input and 2$\times$ the size of the labels (to account for model outputs) divided by the batch size.}
\label{tab:memory_breakdown}
\end{table}

\subsection{Experiment Setup}
\label{sec:exp-details}
The exact hardware configurations are listed below. Software versions are listed in~\cref{tab:benchmark-versions}.

\textbf{End-to-end benchmarks.}
The end-to-end benchmarks were executed within a virtual environment on a public cloud with an Intel(R) Xeon(R) CPU @ 2.20GHz, NVIDIA A100 SXM4 (40GB VRAM), and 83GB RAM. We created and ran a separate Docker container for each framework. The Docker image source is 
nvidia/cuda:11.4.2-cudnn8-devel-ubuntu20.04.

\textbf{Microbenchmarks.} The microbenchmarks were executed on a cloud instance running Ubuntu 18.04.5 LTS with an Intel(R) Xeon(R) Platinum 8275CL CPU @ 3.00GHz, NVIDIA A100 SXM4 (40GB VRAM), and 1.1TB of RAM. CUDA memory was allocated in block sizes of 512.

For details on the settings for each layer, refer to \url{https://github.com/pytorch/opacus/blob/main/benchmarks/config.json}.

\begin{table}
\begin{center}
\begin{tabular}{l| l} 
    \toprule
     \textbf{Software} & \textbf{Version}\\
     \midrule
     \multirow{2}{*}{Python} & 3.8.10 (end-to-end)\\ 
     & 3.9.7 (microbenchmarks)\\
     \midrule
     dm-haiku & 0.0.5\\
     JAX & 0.2.25\\
     jaxlib & 0.1.73 \\
     \midrule
     PyTorch & 1.10.0  \\
     \midrule
     Opacus & 1.0.0 \\
     \midrule
     BackPACK & 0.1\\
     backpack-for-pytorch & 1.4.0\\
     \midrule
     TensorFlow & 2.7.0 \\
     TensorFlow Privacy & 0.7.3\\
     \midrule
     PyVacy & 0.0.1 + commit \texttt{2e0a9f}\\
     \bottomrule
\end{tabular}
\end{center}
\caption{Software versions used in the end-to-end and microbenchmarks. The latter only use Python, PyTorch, and Opacus.}
\label{tab:benchmark-versions}
\end{table}

\end{document}